\documentclass[lettersize,journal]{IEEEtran}
\usepackage{amsmath,amsfonts}
\usepackage{algorithmic}
\usepackage{algorithm}
\usepackage{array}
\usepackage[caption=false,font=normalsize,labelfont=sf,textfont=sf]{subfig}
\usepackage{textcomp}
\usepackage{stfloats}
\usepackage{url}
\usepackage{verbatim}
\usepackage{graphicx}
\usepackage{multirow}
\usepackage[table,xcdraw]{xcolor}
\usepackage{cite}
\begin{document}

\title{Physics-Guided Attention in a Lightweight TCN for Efficient WiFi CSI-Based Human Activity Recognition}

\author{Chinthaka Ranasingha,
        Tharindu~Fernando,~\IEEEmembership{Member,~IEEE,}
        Sridha~Sridharan,~\IEEEmembership{Life Senior Member,~IEEE,}
        Clinton~Fookes,~\IEEEmembership{Senior Member,~IEEE}
        ~and~ Harshala Gammulle, ~\IEEEmembership{Member,~IEEE.}
\thanks{The authors are with the Signal Processing, Artificial Intelligence
and Vision Technologies (SAIVT) Research Group, School of Electrical
Engineering and Robotics, Queensland University of Technology (QUT),
Brisbane, QLD 4000, Australia (e-mail: c.ranasingha@hdr.qut.edu.au).}
\thanks{The research presented in this paper was supported partly by an Australian Research Council (ARC) Discovery grant DP200101942.}}
\markboth{Journal of \LaTeX\ Class Files,~Vol.~14, No.~8, August~2021}
{Shell \MakeLowercase{\textit{et al.}}: A Sample Article Using IEEEtran.cls for IEEE Journals}


\maketitle

\begin{abstract}
Human Action Recognition (HAR) using WiFi Channel State Information (CSI) has gained increasing attention due to its non-contact, low-cost, and privacy-preserving nature. However, existing learning-based approaches largely rely on deep, computationally intensive architectures to implicitly capture motion dynamics from CSI measurements, thereby increasing model complexity and reducing efficiency. 

Instead we argue that incorporating appropriate inductive biases tailored to the physical characteristics of CSI signals enables more efficient and effective learning.

In this work, we propose a compact temporal convolutional network (TCN)-based framework that explicitly incorporates motion-aware inductive biases into feature learning. Specifically, we introduce a Doppler-energy-guided temporal attention mechanism in feature space to emphasize motion-salient time segments, and a variance-driven channel attention module to weight informative subcarriers based on temporal motion statistics adaptively. 

By integrating these domain-specific priors, the proposed model effectively captures motion dynamics without increasing architectural depth. Extensive experiments on multiple benchmark datasets demonstrate that our approach achieves superior performance compared to deeper baselines, while significantly reducing parameter count and computational cost.
\end{abstract}


\section{Introduction}


Human Action Recognition (HAR) is the process of identifying human movement and behavior. Various sensors have been utilized to capture these behaviors \cite{miao2025wi}. WiFi Channel State Information (CSI)-based HAR is gaining tremendous attention compared to uncomfortable wearable sensors and privacy-intrusive cameras \cite{wei2025survey}. Moreover, due to the cost-effective commercial WiFi devices, non-contact privacy-preserving nature of the recognition, and the ability to operate effectively in Non-Line-of-Sight (NLoS) environments \cite{abuhoureyah2024wifi}, applications such as health monitoring \cite{khan2017wireless}, security and surveillance, and many other Internet of Things (IOT) have embraced WiFi-based HAR \cite{liang2023liwi}. 


Human motion induces variations in wireless propagation, including amplitude fluctuations, phase changes, and Doppler frequency shifts. These motion-induced effects have been widely explored in the context of HAR. Early works, such as E-eyes, demonstrate that fine-grained CSI information contains sufficient motion information for HAR through statistical analysis of amplitude and phase variations \cite{wang2014eyes}. Subsequent studies further characterized the physical relationship between human movement and wireless channels, showing that different subcarriers exhibit distinct sensitivity to motion \cite{wang2015understanding}. To better capture motion dynamics, several works have explored time-frequency analysis of CSI. Early approaches such as WiDance \cite{qian2017inferring} and WiSee \cite{pu2013whole} showed that Doppler and frequency-domain representations provide explicit descriptions of motion intensity and temporal evolution, enabling more discriminative activity characterization than raw signal measurements \cite{pu2013whole,qian2017inferring}. Later studies incorporated short-time Fourier transform (STFT) representations into learning-based pipelines to enhance recognition performance \cite{niu2024csi}. While effective, these approaches rely on external feature engineering and do not integrate motion structure directly into the representation learning process.

To improve generalization across users and environments, recent research has increasingly adopted deep learning such as convolutional neural networks (CNNs) \cite{alazrai2020end,xiao2020deepseg,ma2018signfi}, recurrent neural networks (RNNs) \cite{ding2019wifi,cao2019contactless,wang2018channel,chen2018wifi}, and hybrid CNN-RNN architectures \cite{shalaby2022utilizing,mekruksavanich2023attention,zou2018deepsense}. More recently, transformer-based models and multi-branch deep networks have been proposed to further improve recognition accuracy by increasing model depth and representational capacity \cite{hussain2024wisigpro,tan2025wifi,luo2024vision,li2021two,shang2023recurrent,yi2024probsparse}.


Despite their success, state-of-the-art (SOTA) deep learning based WiFi HAR models come at the cost of significant model complexity, leading to high computational overhead and memory consumption. Such designs limit deployment on resource-constrained devices and real-time sensing platforms. Moreover, increasing model depth alone does not explicitly encode the physical characteristics of human motion in wireless signals, forcing networks to implicitly learn motion saliency from raw or weakly processed CSI, which can be inefficient and unstable, especially with limited training data.

In this work, we argue that high recognition performance does not necessarily require deep or computationally expensive models. We demonstrate that incorporating explicit, domain-specific motion biases into a compact model can yield superior performance to simply increasing network depth. Specifically, we propose a lightweight temporal convolutional network (TCN)\cite{lea2017temporal} augmented with two complementary motion-guided attention mechanisms: Doppler-energy-based temporal attention and variance-driven channel attention. Rather than operating directly on raw CSI measurements, both attentions are computed in the learned feature space produced by an initial temporal encoder. This design choice allows the model first to suppress static components and hardware-induced noise, making subsequent motion saliency estimation more robust and semantically meaningful.

The Doppler-energy temporal attention explicitly highlights time segments that exhibit strong motion-induced frequency shifts, enabling the model to focus on action-relevant temporal regions. In parallel, the variance-driven channel attention leverages temporal variance statistics to identify informative subcarriers that are more sensitive to human motion. By jointly modeling temporal and channel-wise motion saliency, the proposed approach effectively captures discriminative activity patterns without increasing network depth or width. Importantly, the introduced attention modules add only minimal computational overhead compared to stacking additional convolutional or transformer layers.

Extensive experiments demonstrate that the proposed model consistently outperforms several deep and parameter-heavy baseline architectures while using an order of magnitude fewer parameters and floating-point operations (FLOPs). These results highlight that carefully designed, physics- and statistics-inspired attention mechanisms can significantly improve recognition accuracy while maintaining computational efficiency. Our findings suggest that incorporating explicit motion priors into lightweight architectures is a promising direction for practical and scalable WiFi-based sensing systems.

The main contributions of this paper are summarized as follows:

\begin{itemize}
    \item We propose a compact TCN-based HAR framework that achieves high accuracy without relying on deep or computationally expensive architectures.
    \item We introduce feature-space Doppler-energy temporal attention to emphasize motion-salient time segments in CSI-derived features explicitly.
    \item We design a variance-driven channel attention mechanism that highlights informative subcarriers based on temporal motion statistics.
    \item We demonstrate that the proposed approach outperforms deeper baselines while significantly reducing parameter count and computational cost through extensive experiments.
\end{itemize}

\section{Related Works}
\label{sec:related}

\subsection{Deep Learning for CSI-based Human Action Recognition}

Recent advances in WiFi CSI-based Human Action Recognition (HAR) have been predominantly driven by deep learning models that learn discriminative representations directly from raw or minimally processed CSI measurements. CNNs are widely used for transforming CSI streams into image-like representations, thereby enabling hierarchical feature extraction via convolutional filters. For example, Moshiri et al. \cite{moshiri2021csi} convert CSI signals into pseudo-colored images and employ stacked 2D convolutions for action recognition, while subsequent works extend CNN-based designs to jointly address tasks such as action segmentation, classification, and localization \cite{yan2021device,xiao2020deepseg}.

Recurrent architectures further model CSI as temporal sequences to capture time-dependent dynamics. Variants such as Long Short-Term Memory (LSTM), Bidirectional LSTM (BLSTM), and more recent sequence models such as Mamba \cite{gu2024mamba} have been employed to encode temporal dependencies. Several works incorporate signal preprocessing techniques, such as Discrete Wavelet Transform (DWT) or statistical feature extraction, to improve robustness prior to sequence modeling \cite{ding2019wifi, chen2018wifi, tan2025wifi}. While these approaches enhance temporal modeling, they still rely on the network to learn motion characteristics from transformed representations.

More recently, Transformer-based architectures have been introduced to capture long-range dependencies through self-attention mechanisms. These models leverage multi-head attention and positional encoding to model complex spatio-temporal interactions in CSI data \cite{hussain2024wisigpro, luo2024vision,yi2024probsparse}. Hybrid architectures combining CNNs with Transformers have also been explored to jointly encode local and global features \cite{li2021two,shang2023recurrent}.

In parallel, a subset of works attempts to incorporate motion-related information through signal transformations, such as Doppler spectrograms or time–frequency representations derived via Short-Time Fourier Transform (STFT) or DWT \cite{yang2023witransformer,wang2018channel,showmik2023human}. While these approaches introduce domain knowledge at the preprocessing stage, the extracted representations are typically treated as generic inputs to deep models, without explicitly embedding motion-aware structure within the learning architecture itself.

Despite architectural diversity, these methods share a common underlying paradigm: they rely on increasing model capacity to implicitly learn motion dynamics from CSI data in an end-to-end manner. In this setting, motion-related patterns arising from physical phenomena such as Doppler shifts and multipath propagation are not explicitly modeled, but are expected to emerge through deep hierarchical feature extraction, which will result in over-parameterized architectures, leading to higher computational cost and a lack of interpretability of learned representation.

These observations suggest that the prevailing trend of scaling model complexity is not well aligned with the underlying structure of CSI signals, motivating the need for approaches that explicitly integrate domain-specific motion priors into compact and efficient learning frameworks.

\subsection{Attention Mechanisms and Feature Reweighting in HAR}

Attention mechanisms have been widely adopted in HAR to enhance feature representation by adaptively reweighting informative components. In CSI-based HAR, temporal attention mechanisms are employed to emphasize salient time segments, while channel attention modules selectively weight subcarriers or antennas based on their contribution to recognition performance. Additionally, spatio-temporal attention and self-attention mechanisms, particularly in Transformer-based architectures, have been introduced to model long-range dependencies and complex interactions within CSI signals \cite{hussain2024wisigpro, li2021two, shang2023recurrent, yi2024probsparse}. Earlier works have also incorporated attention into recurrent frameworks, enabling models to dynamically focus on informative time steps and features during sequence modeling \cite{chen2018wifi}.

Despite their effectiveness, existing attention mechanisms are predominantly data-driven, learning feature importance solely from supervision signals without explicit consideration of the underlying physical properties of CSI measurements. Moreover, because attention is learned implicitly, it does not fundamentally reduce the reliance on deep architectures, and it further increases parameters and computational overhead. Consequently, while attention mechanisms improve performance, they do not address the absence of explicit motion-aware modeling within the network.

In contrast, this work departs from purely data-driven attention by introducing motion-aware, feature-space attention mechanisms that explicitly encode domain-specific priors within a compact architecture.

\subsection{Inductive Bias and Efficient Modeling for Signal-based HAR}

Other than purely data-driven approaches, another line of work has explored the use of domain knowledge to explicitly model motion dynamics in CSI-based HAR. Early methods rely on handcrafted features derived from signal processing principles, such as Doppler shifts, signal energy, variance, and correlation statistics, to characterize human motion from raw data. These approaches often transform CSI measurements into time–frequency representations using techniques such as Discrete Wavelet Transform (DWT) or Short-Time Fourier Transform (STFT), followed by conventional classifiers or shallow learning models \cite{zou2017multiple, guo2019wiar, huang2021phaseanti, ding2019wifi, showmik2023human}. While these methods provide interpretable and physically meaningful representations, they are typically constrained by fixed feature extraction pipelines, limiting their adaptability to complex and diverse activity patterns.

In parallel, recent efforts have focused on improving computational efficiency through lightweight model designs. These approaches aim to lower computational cost and enable real-time deployment in resource-constrained environments. However, efficiency in such methods is primarily achieved by reducing model capacity rather than improving motion representation, and they continue to rely on implicit learning of motion dynamics from CSI data \cite{zhang2025mkfi, sai2026machine}. Consequently, performance often degrades when model complexity is constrained.

Above, explicit signal-driven modeling and efficient deep learning highlight a fundamental trade-off in existing CSI-based HAR systems. Methods grounded in domain knowledge introduce strong inductive bias but lack flexibility, whereas deep learning approaches provide adaptability at the cost of increased complexity and weak incorporation of physical priors, which motivates the need for systems that incorporate motion-aware inductive biases directly into compact, learnable architectures.

\section{Proposed Method}
\label{sec:modeldesign}

Channel state information (CSI) describes how the signal propagates from the transmitter to the receiver and represents the combined effects of scattering, fading, and attenuation in MIMO-OFDM (Multiple Input Multiple Output Orthogonal Frequency Division Multiplexing) systems (e.g., commercial WiFi devices compliant with IEEE 802.11n). For each transmitter–receiver antenna pair and subcarrier
$f$ at time $t$, the frequency-domain relationship between transmitted and received signals can be expressed as:
\begin{equation}
    Y(f,t) = H(f,t)X(f,t) + \epsilon ,
\end{equation}

where $X(f,t)$ and $Y(f,t)$ denote the transmitted and received signals, respectively, $H(f,t)$ represents the complex channel response, and $\epsilon$ models measurement noise and interference. In WiFi sensing systems, CSI provides estimates of the complex channel response $H(f,t)$ across multiple subcarriers over time.

The complex CSI can be represented in magnitude–phase form as:
\begin{equation}
    H(f,t) = |H(f,t)|e^{i\phi(f,t)},
\end{equation}

where $|H(f,t)|$ and $\phi(f,t)$ denote the amplitude and phase of the channel response, respectively. Although amplitude responses provide relatively stable motion-related energy variations, discriminative motion structure is not directly observable from raw CSI due to environmental interference and measurement instability.



\begin{figure*}[t]
    \centering
    \includegraphics[width=0.94\textwidth]{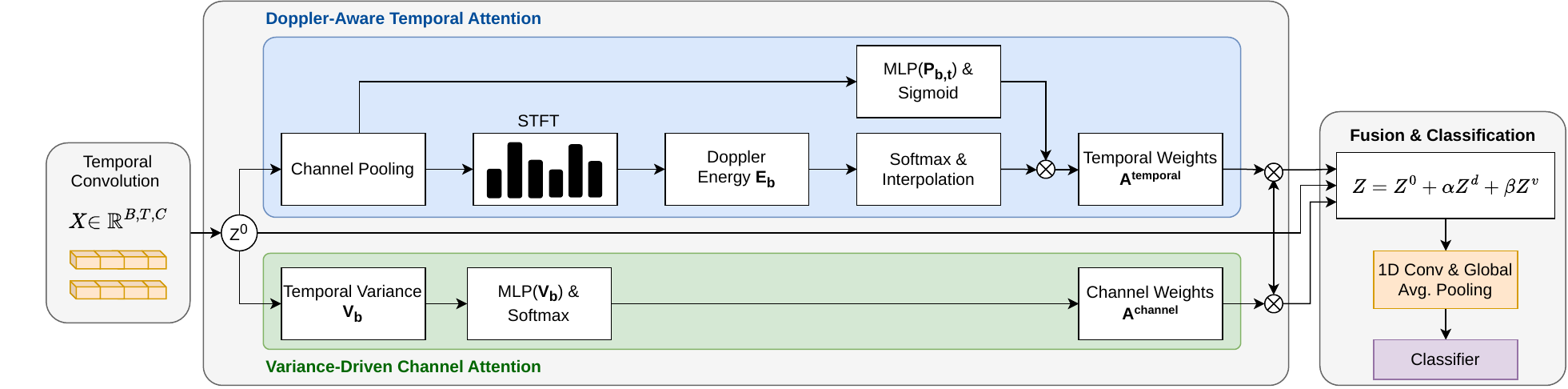}
    \caption{Overview of the proposed method.}
    \label{fig:overview}
\end{figure*}

To model motion dynamics from WiFi CSI measurements, we consider the CSI amplitude observations collected over $T$ time steps and $C$ flattened CSI feature channels. ( $C=$ subcarriers $\times$ antennas)
\begin{equation}
    X = |H(f,t)| \in \mathbb{R}^{B\times T\times C},
\end{equation}
where $B$ is the batch dimension.

As illustrated in Fig. \ref{fig:overview}, our proposed framework consists of three main components: (i) a temporal convolutional encoder that learns noise-robust motion representations from CSI amplitude sequences, (ii) motion-guided feature weighting modules that emphasize discriminative temporal and spectral patterns, and (iii) a learnable fusion and classification module. The encoder captures local temporal dependencies and suppresses static channel components, producing a compact feature representation. On top of this representation, complementary temporal and channel-wise weighting mechanisms are applied to enhance motion-salient dynamics without increasing model depth. Finally, a learnable fusion mechanism fuses the learned representations with the original features via a residual connection. The design of each component is described in detail in the following subsections.

\subsection{Temporal Convolutional Encoder}


Raw CSI sequences are noisy by nature. We employ TCN blocks, composed of stacked 1D dilated convolutions, to capture temporal dependencies in CSI signals. The input tensor is permuted and processed with two sequential TCN layers, each followed by batch normalization, ReLU activation, and dropout.

\begin{multline}
    H^1 = \text{Drop(ReLU(BN(Conv}_{d=1}^{r}(X)))), \\ 
    Z^0_{b,t,c} = \text{Drop(ReLU(BN(Conv}_{d=2}^{r}(H^1))))
    \label{tcn_encoder}
\end{multline}

where $Conv_d(\cdot)$ denotes dilated convolution with dilation factor $d$. We kept $d=1$ for the first layer and $d=2$ for the second. The first layer applies a channel reduction with factor $r$, compressing the feature dimension from $C$ to $\frac{C}{r}$, while the second layer restores it back to $C$. This bottleneck design reduces computational cost while preserving essential temporal features.

\subsection{Doppler-energy Attention}

Human motion induces frequency shifts in wireless signals. To exploit these properties, we use a Doppler-energy-based attention mechanism to highlight the motion-heavy time regions. This mainly integrates Short-Time Fourier Transform (STFT)-based Doppler energy with learnable temporal attention.

We perform mean pooling across the feature dimension to obtain a compact motion-sensitive signal in the temporal domain while reducing redundant information:
\begin{equation}
    P_{b,t}
    =
    \frac{1}{C}
    \sum_{c=1}^{C}
    Z^0_{b,t,c}.
\end{equation}

This pooled signal preserves temporal motion patterns while suppressing feature-level noise.

We compute the Short-Time Fourier Transform (STFT) of the pooled signal to capture local frequency variations:
\begin{equation}
    S_b(f,k)
    =
    \sum_{t}
    P_{b,t}\,
    w(t-kH)\,
    e^{-j2\pi ft/N},
    \label{eq:fourier}
\end{equation}

where $P_{b,t}$ denotes the pooled CSI signal at time index $t$, $N$ is the analysis window length, $H$ is the hop size, and $w(\cdot)$ is a Hann window function. Here, $f$ indexes discrete frequency bins and $k$ indexes time-frequency segments determined by the sliding window. The Hann window is adopted to reduce spectral leakage and improve time-frequency localization for Doppler-sensitive motion analysis.



Then we obtain the spectral energy representation.
\begin{equation}
    M_b(f,k) = |S_b(f,k)|^2.
\end{equation}

To derive a compact representation of motion intensity, we aggregate energy across frequency bins:
\begin{equation}
    E_b(k) = \sum_f M_b(f,k).
\end{equation}

This aggregation captures overall motion-induced energy regardless of the exact Doppler frequency, providing a robust measure of temporal motion saliency.

The Doppler energy is normalized using a temperature-scaled softmax:
\begin{equation}
    {A}_b(k) = \frac{\exp(E_b(k)/\tau)}{\sum_{k'} \exp(E_b(k')/\tau)}
    \label{softmax_dopler}
\end{equation}

where $\tau > 0$ is the temperature parameter controlling the sharpness of the attention distribution, $k$ indexes the STFT frames.

Since STFT produces a reduced temporal resolution ($K<T$), we upsample the Doppler attention to match the original sequence length:
\begin{equation}
    \hat{A}_{b,t} = I({A_b}(k), T), \quad t = 1, \ldots, T 
\end{equation}

where $I(\cdot,T)$ denotes 1D linear interpolation along the temporal dimension.

In parallel, a lightweight temporal bottleneck MLP compresses the temporal descriptor from $T$ to $\frac{T}{r_A}$ and restores it to $T$.

\begin{equation}
    A_{b}^{MLP}= \sigma(W_2\, ReLU(W_1\, P_{b})),
    \label{mlp_dopler}
\end{equation}

where  $P_{b}\in \mathbb{R}^T$ and $W_1 \in \mathbb{R}^{\frac{T}{r_A}\times T}, W_2 \in \mathbb{R}^{T \times \frac{T}{r_A}}$ are weight matrices. $\sigma(\cdot)$ denotes the sigmoid activation.

To combine physics-driven and learnable attention, we employ element-wise fusion:
\begin{equation}
    \widetilde{A}_{b,t} = \hat{A}_{b,t} \cdot A_{b}^{MLP},
\end{equation}
where $A_{b}^{MLP} \in \mathbb{R}^T$.
This design emphasizes frames with strong motion-induced Doppler energy while retaining adaptability to dataset-specific patterns.

We further normalized attention weights to maintain numerical stability and consistent scaling across samples:
\begin{equation}
    A^{temporal}_{b,t} = \frac{\widetilde{A}_{b,t}}{\sum_{t'=1}^{T} \widetilde{A}_{b,t'} + \epsilon}
\end{equation}

Finally, the normalized attention weights are applied to the CSI tensor.
\begin{equation}
    Z_{b,t,c}^d =  Z^0_{b,t,c} \cdot A^{temporal}_{b,t}
\end{equation}

\subsection{Variance-based Channel Attention}

Human motion-induced perturbations in wireless channels result in CSI amplitude fluctuations across subcarriers. Subcarrier paths that make direct contact may exhibit higher variation, while static or noise-dominated subcarriers show relatively stable responses.

To quantify motion sensitivity, we compute the temporal variance for each flattened feature dimension of the encoded CSI features over $T$ time steps:
\begin{equation}
    V_{b,c} = \frac{1}{T} \sum_{t=1}^{T} \left(Z^0_{b,t,c} - \mu_{b,c}\right)^2, \text{where,}
\end{equation}
\begin{equation}
\mu_{b,c} = \frac{1}{T} \sum_{t=1}^{T} Z^0_{b,t,c}.
\end{equation}
This operation aggregates temporal fluctuations and produces a compact representation that highlights dynamically responsive subcarriers. This variance descriptor is then passed through a lightweight bottleneck MLP with a reduction ratio $r_V$.
\begin{equation}
    l_{b}= W_2\, ReLU(W_1\, V_{b}),
    \label{mlp_variance}
\end{equation}

where $W_1 \in \mathbb{R}^{\frac{C}{r_V}\times C}, W_2 \in \mathbb{R}^{C \times \frac{C}{r_V}}$.
We then normalized the channel attention weights using a temperature-scaled softmax, with temperature parameter $\tau$.
\begin{equation}
    A^{channel}_{b,c} = \frac{\exp(l_{b,c}/\tau)}{\sum_{c'=1}^{C} \exp(l_{b,c'}/\tau)}
    \label{softmax_variance}
\end{equation}

Finally, we re-weight the original encoded CSI features using the computed channel attention weights, thereby enhancing motion-relevant subcarriers while suppressing noise-prone or less informative channels.
\begin{equation}
    Z_{b,t,c}^v =  Z^0_{b,t,c} \cdot A^{channel}_{b,c}.
\end{equation}

\subsection{Dual Attention Fusion and Classification}

We fuse temporal and channel attentions via a learnable modulation, with $\alpha$ and $\beta$ trainable scalar parameters that adaptively balance Doppler-temporal and variance-channel contributions. This formulation preserves the original representation while allowing dynamic refinement.
\begin{equation}
    Z_b =  Z^0 + \alpha Z^d + \beta Z^v
\end{equation}

Then, the features are further refined using an additional lightweight temporal convolutional block, followed by global average pooling, before linear classification.
\begin{equation}
    \hat{Z} = \text{Drop(ReLU(Conv}_d^r(Z_b)))
\end{equation}

\section{Experiments and Analysis}

\subsection{Datasets}

We evaluate our proposed model over three public datasets. The CSI-HAR dataset \cite{moshiri2021csi} contains seven different human activities performed by three individuals. Each activity was performed 20 times, resulting in 420 samples. The authors initially split the dataset into training and test sets in a $75\%$ to $25\%$ ratio. Later, Varga \cite{varga2024mitigating} proposed a split in which the training set comprises the first two subjects, while the last subject is in the testing set, thereby mitigating the data leakage issue. We use the newly proposed split in our experiments. The NTU-Fi HAR \cite{yang2023sensefi} dataset comprises six action classes and contains $1200$ samples in total. Each sample has dimension $342$, corresponding to $114$ subcarriers across 3 antenna pairs. Following previous work \cite{li2025human,tan2025wifi}, we use the provided split, with $936$ samples in the training set and $264$ in the test set. The UT-HAR dataset \cite{yousefi2017survey} has seven action classes and contains $3977$ and $996$ samples in the training and testing sets, respectively.

\subsection{Hyperparameter and Experiment Settings}

We used PyTorch version 2.0.1. CSI-HAR: We used the Adam optimizer with an initial learning rate of $0.001$, hyperparameters $\beta_1=0.9$ and $\beta_2=0.99$, and a weight decay of $1e^{-4}$. The fine-tuned reduction factors in Eq. \ref{tcn_encoder},\ref{mlp_dopler},\ref{mlp_variance} were set to $2,1,4$, respectively and is trained for $150$ epochs. The temperature parameters of Eq. \ref{softmax_dopler},\ref{softmax_variance} were set to $1,0.3$ respectively and the hyperparameter window length (N) and hop size (H) of equation \ref{eq:fourier} are $50$ and $40$ respectively. NTU-FI HAR: We used Adam optimizer with default parameters with reduction factors $18,4,1$ for Eq. \ref{tcn_encoder},\ref{mlp_dopler},\ref{mlp_variance} respectively, and both temperature parameters of Eq. \ref{softmax_dopler},\ref{softmax_variance} were set to $1$. This was trained for $50$ epochs with the hyperparameter window length (N) and hop size (H) of equation \ref{eq:fourier} as $50$ and $10$ respectively. UT-HAR: The Adam optimizer is used with default parameters, and the factors in Eqs. \ref{tcn_encoder}, \ref{mlp_dopler}, and \ref{mlp_variance} are $ 1, 12, 3$, respectively. Both Temperature parameters in Eqs. \ref{softmax_dopler} and \ref{softmax_variance} were set to $1$ and trained for $100$ epochs. The hyperparameter window length (N) and hop size (H) of equation \ref{eq:fourier} for this dataset were set to $10$ and $5$ respectively.

\subsection{Performance Comparison and Quantitative Results }

\begin{table*}[htpb]
\centering
\begin{tabular}{|
>{\columncolor[HTML]{ECF4FF}}l |
>{\columncolor[HTML]{DAE8FC}}c 
>{\columncolor[HTML]{DAE8FC}}c 
>{\columncolor[HTML]{DAE8FC}}c |
>{\columncolor[HTML]{ECF4FF}}c 
>{\columncolor[HTML]{ECF4FF}}c 
>{\columncolor[HTML]{ECF4FF}}c |
>{\columncolor[HTML]{DAE8FC}}c 
>{\columncolor[HTML]{DAE8FC}}c 
>{\columncolor[HTML]{DAE8FC}}c |}
\hline
\multicolumn{1}{|c|}{\cellcolor[HTML]{ECF4FF}} &
  \multicolumn{3}{c|}{\cellcolor[HTML]{DAE8FC}UT-HAR} &
  \multicolumn{3}{c|}{\cellcolor[HTML]{ECF4FF}NTU-Fi-HAR} &
  \multicolumn{3}{c|}{\cellcolor[HTML]{DAE8FC}CSI-HAR} \\ \cline{2-10} 
\multicolumn{1}{|c|}{\multirow{-2}{*}{\cellcolor[HTML]{ECF4FF}Model}} &
  \multicolumn{1}{c|}{\cellcolor[HTML]{DAE8FC}\begin{tabular}[c]{@{}c@{}}Accuracy \\ (\%)\end{tabular}} &
  \multicolumn{1}{c|}{\cellcolor[HTML]{DAE8FC}\begin{tabular}[c]{@{}c@{}}FLOPS \\ (M)\end{tabular}} &
  \begin{tabular}[c]{@{}c@{}}Parameters \\ (M)\end{tabular} &
  \multicolumn{1}{c|}{\cellcolor[HTML]{ECF4FF}\begin{tabular}[c]{@{}c@{}}Accuracy \\ (\%)\end{tabular}} &
  \multicolumn{1}{c|}{\cellcolor[HTML]{ECF4FF}\begin{tabular}[c]{@{}c@{}}FLOPS \\ (M)\end{tabular}} &
  \begin{tabular}[c]{@{}c@{}}Parameters \\ (M)\end{tabular} &
  \multicolumn{1}{c|}{\cellcolor[HTML]{DAE8FC}\begin{tabular}[c]{@{}c@{}}Accuracy \\ (\%)\end{tabular}} &
  \multicolumn{1}{c|}{\cellcolor[HTML]{DAE8FC}\begin{tabular}[c]{@{}c@{}}FLOPS \\ (M)\end{tabular}} &
  \begin{tabular}[c]{@{}c@{}}Parameters \\ (M)\end{tabular} \\ \hline
ResNet18* &
  \multicolumn{1}{c|}{\cellcolor[HTML]{DAE8FC}98.11} &
  \multicolumn{1}{c|}{\cellcolor[HTML]{DAE8FC}49.93} &
  11.180 &
  \multicolumn{1}{c|}{\cellcolor[HTML]{ECF4FF}95.31} &
  \multicolumn{1}{c|}{\cellcolor[HTML]{ECF4FF}54.19} &
  11.180 &
  \multicolumn{1}{c|}{\cellcolor[HTML]{DAE8FC}54.59} &
  \multicolumn{1}{c|}{\cellcolor[HTML]{DAE8FC}26.404} &
  11.183 \\ \hline
ResNet50* &
  \multicolumn{1}{c|}{\cellcolor[HTML]{DAE8FC}97.21} &
  \multicolumn{1}{c|}{\cellcolor[HTML]{DAE8FC}86.40} &
  23.550 &
  \multicolumn{1}{c|}{\cellcolor[HTML]{ECF4FF}99.38} &
  \multicolumn{1}{c|}{\cellcolor[HTML]{ECF4FF}90.66} &
  23.550 &
  \multicolumn{1}{c|}{\cellcolor[HTML]{DAE8FC}48.99} &
  \multicolumn{1}{c|}{\cellcolor[HTML]{DAE8FC}48.029} &
  23.551 \\ \hline
RNN* &
  \multicolumn{1}{c|}{\cellcolor[HTML]{DAE8FC}83.53} &
  \multicolumn{1}{c|}{\cellcolor[HTML]{DAE8FC}\textbf{2.51}} &
  \textbf{0.010} &
  \multicolumn{1}{c|}{\cellcolor[HTML]{ECF4FF}84.64} &
  \multicolumn{1}{c|}{\cellcolor[HTML]{ECF4FF}\textbf{13.09}} &
  \textbf{0.027} &
  \multicolumn{1}{c|}{\cellcolor[HTML]{DAE8FC}30.54} &
  \multicolumn{1}{c|}{\cellcolor[HTML]{DAE8FC}\textbf{3.808}} &
  \textbf{0.008} \\ \hline
GRU* &
  \multicolumn{1}{c|}{\cellcolor[HTML]{DAE8FC}94.18} &
  \multicolumn{1}{c|}{\cellcolor[HTML]{DAE8FC}7.60} &
  0.030 &
  \multicolumn{1}{c|}{\cellcolor[HTML]{ECF4FF}97.66} &
  \multicolumn{1}{c|}{\cellcolor[HTML]{ECF4FF}39.39} &
  0.079 &
  \multicolumn{1}{c|}{\cellcolor[HTML]{DAE8FC}62.42} &
  \multicolumn{1}{c|}{\cellcolor[HTML]{DAE8FC}11.552} &
  0.023 \\ \hline
LSTM* &
  \multicolumn{1}{c|}{\cellcolor[HTML]{DAE8FC}87.18} &
  \multicolumn{1}{c|}{\cellcolor[HTML]{DAE8FC}10.14} &
  0.040 &
  \multicolumn{1}{c|}{\cellcolor[HTML]{ECF4FF}97.14} &
  \multicolumn{1}{c|}{\cellcolor[HTML]{ECF4FF}52.54} &
  0.105 &
  \multicolumn{1}{c|}{\cellcolor[HTML]{DAE8FC}44.97} &
  \multicolumn{1}{c|}{\cellcolor[HTML]{DAE8FC}15.360} &
  0.031 \\ \hline
BiLSTM* &
  \multicolumn{1}{c|}{\cellcolor[HTML]{DAE8FC}90.19} &
  \multicolumn{1}{c|}{\cellcolor[HTML]{DAE8FC}20.29} &
  0.080 &
  \multicolumn{1}{c|}{\cellcolor[HTML]{ECF4FF}99.69} &
  \multicolumn{1}{c|}{\cellcolor[HTML]{ECF4FF}105.09} &
  0.209 &
  \multicolumn{1}{c|}{\cellcolor[HTML]{DAE8FC}40.94} &
  \multicolumn{1}{c|}{\cellcolor[HTML]{DAE8FC}30.720} &
  0.061 \\ \hline
ViT* &
  \multicolumn{1}{c|}{\cellcolor[HTML]{DAE8FC}96.53} &
  \multicolumn{1}{c|}{\cellcolor[HTML]{DAE8FC}273.10} &
  10.580 &
  \multicolumn{1}{c|}{\cellcolor[HTML]{ECF4FF}93.75} &
  \multicolumn{1}{c|}{\cellcolor[HTML]{ECF4FF}501.64} &
  1.052 &
  \multicolumn{1}{c|}{\cellcolor[HTML]{DAE8FC}53.69} &
  \multicolumn{1}{c|}{\cellcolor[HTML]{DAE8FC}66.604} &
  0.663 \\ \hline
Zhang et al. \cite{zhang2025mkfi} &
  \multicolumn{1}{c|}{\cellcolor[HTML]{DAE8FC}98.19} &
  \multicolumn{1}{c|}{\cellcolor[HTML]{DAE8FC}-} &
  0.42 &
  \multicolumn{1}{c|}{\cellcolor[HTML]{ECF4FF}-} &
  \multicolumn{1}{c|}{\cellcolor[HTML]{ECF4FF}-} &
  - &
  \multicolumn{1}{c|}{\cellcolor[HTML]{DAE8FC}-} &
  \multicolumn{1}{c|}{\cellcolor[HTML]{DAE8FC}-} &
  - \\ \hline
Liu et al.\cite{liu2024stackfi} &
  \multicolumn{1}{c|}{\cellcolor[HTML]{DAE8FC}99.13} &
  \multicolumn{1}{c|}{\cellcolor[HTML]{DAE8FC}89.87} &
  12.610 &
  \multicolumn{1}{c|}{\cellcolor[HTML]{ECF4FF}99.27} &
  \multicolumn{1}{c|}{\cellcolor[HTML]{ECF4FF}102.61} &
  11.270 &
  \multicolumn{1}{c|}{\cellcolor[HTML]{DAE8FC}-} &
  \multicolumn{1}{c|}{\cellcolor[HTML]{DAE8FC}-} &
  \cellcolor[HTML]{DAE8FC} - \\ \hline
Yi et al.\cite{yi2024probsparse} &
  \multicolumn{1}{c|}{\cellcolor[HTML]{DAE8FC}-} &
  \multicolumn{1}{c|}{\cellcolor[HTML]{DAE8FC}-} &
  - &
  \multicolumn{1}{c|}{\cellcolor[HTML]{ECF4FF}100.00} &
  \multicolumn{1}{c|}{\cellcolor[HTML]{ECF4FF}553.87} &
  - &
  \multicolumn{1}{c|}{\cellcolor[HTML]{ECF4FF}-} &
  \multicolumn{1}{c|}{\cellcolor[HTML]{ECF4FF}-} &
  \cellcolor[HTML]{ECF4FF}- \\ \hline
Luo et al.\cite{luo2024vision} &
  \multicolumn{1}{c|}{\cellcolor[HTML]{DAE8FC}98.78} &
  \multicolumn{1}{c|}{\cellcolor[HTML]{DAE8FC}-} &
  1.160 &
  \multicolumn{1}{c|}{\cellcolor[HTML]{ECF4FF}98.20} &
  \multicolumn{1}{c|}{\cellcolor[HTML]{ECF4FF}-} &
  2.300 &
  \multicolumn{1}{c|}{\cellcolor[HTML]{CBCEFB}40.00} &
  \multicolumn{1}{c|}{\cellcolor[HTML]{CBCEFB}44.643} &
  \cellcolor[HTML]{CBCEFB}1.645 \\ \hline
Li et al.\cite{li2025human} &
  \multicolumn{1}{c|}{\cellcolor[HTML]{DAE8FC}98.44} &
  \multicolumn{1}{c|}{\cellcolor[HTML]{DAE8FC}70.97} &
  1.760 &
  \multicolumn{1}{c|}{\cellcolor[HTML]{ECF4FF}99.69} &
  \multicolumn{1}{c|}{\cellcolor[HTML]{ECF4FF}75.02} &
  0.570 &
  \multicolumn{1}{c|}{\cellcolor[HTML]{CBCEFB}69.69} &
  \multicolumn{1}{c|}{\cellcolor[HTML]{CBCEFB}1607.559} &
  \cellcolor[HTML]{CBCEFB} 0.996 \\ \hline
Tan et al.\cite{tan2025wifi} &
  \multicolumn{1}{c|}{\cellcolor[HTML]{DAE8FC}99.02} &
  \multicolumn{1}{c|}{\cellcolor[HTML]{DAE8FC}142.32} &
  0.619 &
  \multicolumn{1}{c|}{\cellcolor[HTML]{ECF4FF}100.00} &
  \multicolumn{1}{c|}{\cellcolor[HTML]{ECF4FF}421.77} &
  2.309 &
  \multicolumn{1}{c|}{\cellcolor[HTML]{CBCEFB}72.15} &
  \multicolumn{1}{c|}{\cellcolor[HTML]{CBCEFB}584.418} &
  \cellcolor[HTML]{CBCEFB} 1.098 \\ \hline
OURS &
  \multicolumn{1}{c|}{\cellcolor[HTML]{DAE8FC}\textbf{99.20}} &
  \multicolumn{1}{c|}{\cellcolor[HTML]{DAE8FC}30.68} &
  0.252 &
  \multicolumn{1}{c|}{\cellcolor[HTML]{ECF4FF}\textbf{100.00}} &
  \multicolumn{1}{c|}{\cellcolor[HTML]{ECF4FF}49.82} &
  0.459 &
  \multicolumn{1}{c|}{\cellcolor[HTML]{DAE8FC}\textbf{77.18}} &
  \multicolumn{1}{c|}{\cellcolor[HTML]{DAE8FC}10.798} &
  0.523 \\ \hline
\end{tabular}
\caption{Comparison of performance on the UT-HAR, NTU-Fi HAR, and CSI-HAR datasets. Methods marked with * are reproduced based on the implementation described in \cite{yang2023sensefi}. Results with background color \textcolor{purple!70!black}{purple} are generated by implementing the models of corresponding publications.}
\label{tab:main_table}
\end{table*}

We compare the proposed model with existing WiFi-based HAR methods in terms of accuracy, computational cost, and model capacity. UT-HAR: As shown in Tab. \ref{tab:main_table}, our proposed model outperforms Li et. al. \cite{li2025human} and Tan et. al. \cite{tan2025wifi} by $+0.67\%, +0.18\%$, while reducing FLOPS by $40.29M,111.64M$ and parameters by $1.508M,0.367M$, respectively. On the NTU-Fi HAR dataset, our compact model matches the accuracy of Tan et al. \cite{tan2025wifi} while using only 20\% of the parameters and roughly 12\% of the FLOPs. CSI-HAR: our approach achieves 77.18\% accuracy, outperforming Li et al. \cite{li2025human} by +7.5\% and Tan et al. \cite{tan2025wifi} by +5.0\%, despite using far fewer resources. 

\begin{figure}[h]
    \centering
    \includegraphics[width=0.47\textwidth]{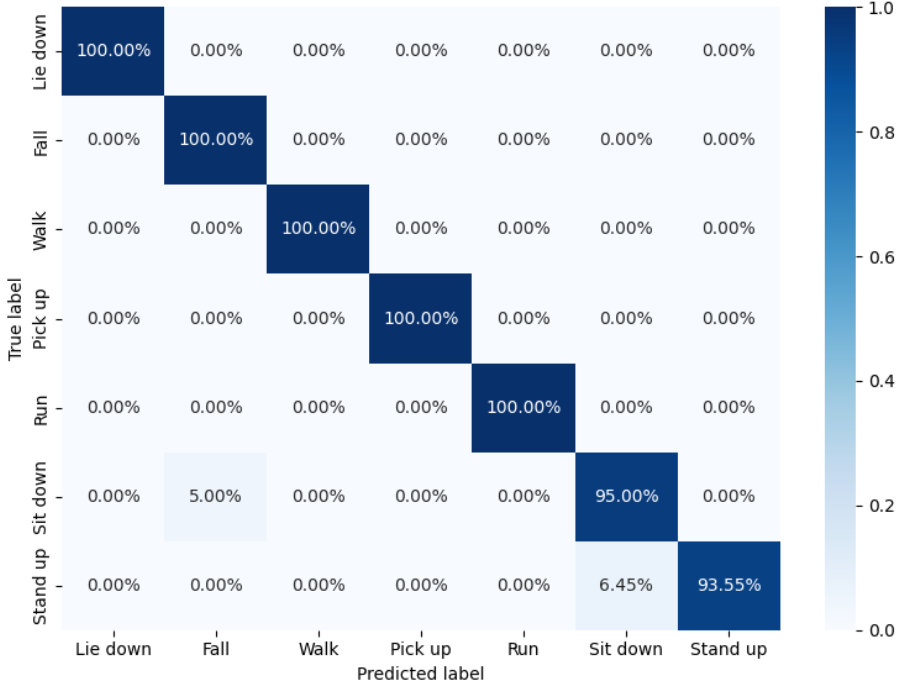}
    \caption{Confusion matrix for UT-HAR dataset}
    \label{fig:con_uthar}
\end{figure}

\begin{figure}[h]
    \centering
    \includegraphics[width=0.47\textwidth]{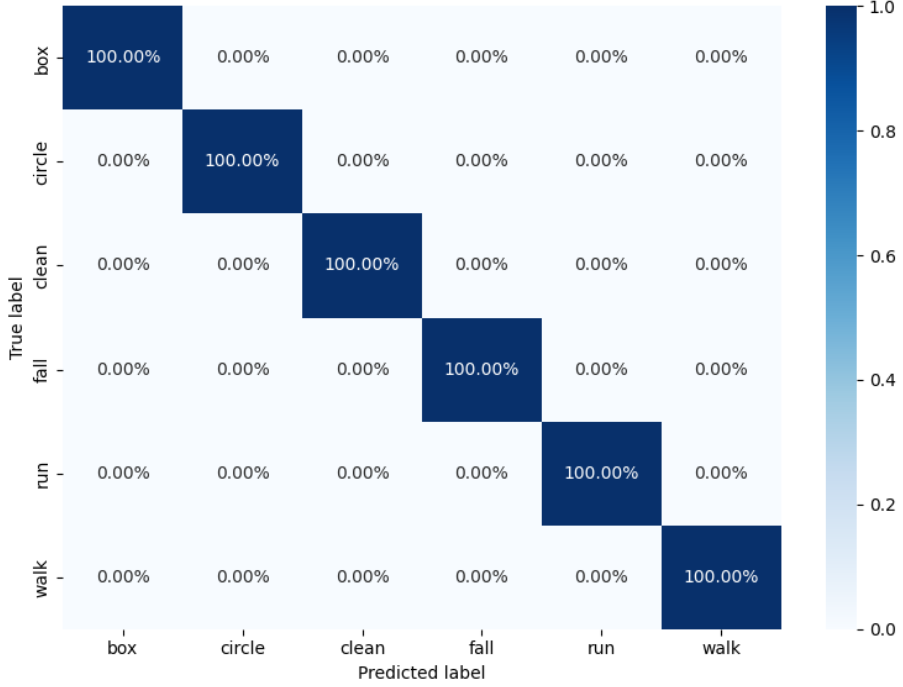}
    \caption{Confusion matrix for NTU-Fi HAR dataset}
    \label{fig:con_ntufihar}
\end{figure}

\begin{figure}[!h]
    \centering
    \includegraphics[width=0.47\textwidth]{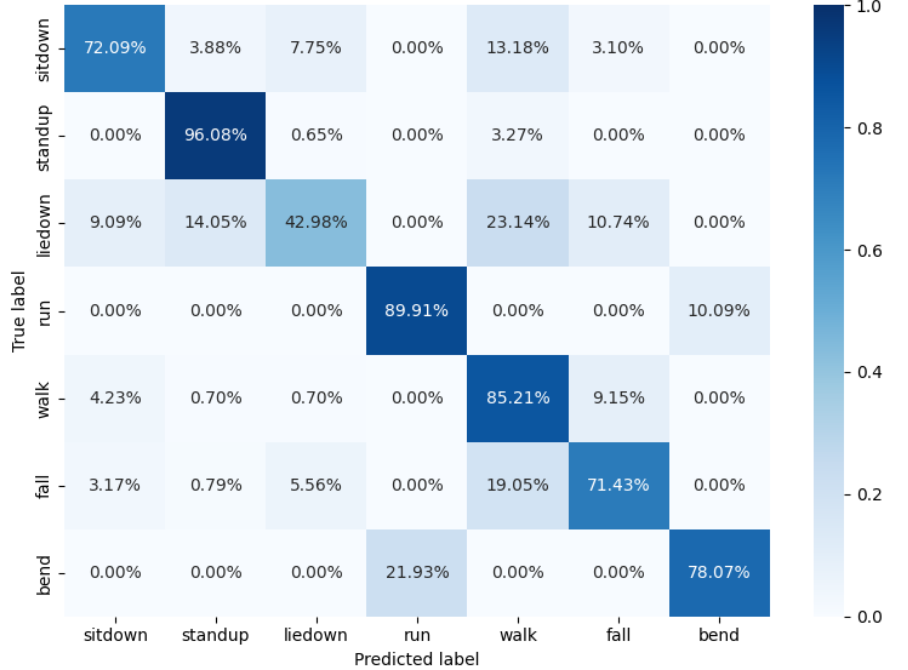}
    \caption{Confusion matrix for CSI-HAR dataset}
    \label{fig:con_csihar}
\end{figure}

Figs. \ref{fig:con_uthar}, \ref{fig:con_ntufihar} and \ref{fig:con_csihar} present the confusion matrices for the UT-HAR, NTU-Fi HAR, and CSI-HAR datasets, respectively. These matrices illustrate the effectiveness of the proposed model, which correctly classifies the majority of actions. For the CSI-HAR dataset, actions with distinctive dynamics, such as stand-up ($96.08\%$), run ($89.91\%$), and walk ($85.21\%$), achieve high classification accuracy. Lower performance is observed for temporally or spatially similar actions, such as lie-down ($42.98\%$) and sit-down ($72.09\%$), which are occasionally misclassified. These results reflect the intrinsic similarity of certain actions, highlighting our model's effectiveness in capturing class-specific motion patterns while minimizing off-diagonal errors.


\subsection{Ablation Experiments and Hyperparameter Tuning}

We conduct ablation experiments on the CSI-HAR dataset to evaluate the contribution of each component. As shown in Tab. \ref{tab:ablation}, the Baseline model, which consists only of TCN blocks, achieves an accuracy of $70.25\%$. Incorporating variance-based channel attention (Baseline+Var) improves the performance to $72.15\%$. Similarly, integrating Doppler-aware attention (Baseline+Doppler) increases the accuracy to $72.82\%$. When both variance-based and Doppler-aware attention mechanisms are applied, the full model achieves $77.18\%$, demonstrating the complementary benefits of the two components.

\begin{table}[!t]
\centering
\begin{tabular}{|
>{\columncolor[HTML]{DAE8FC}}l |
>{\columncolor[HTML]{ECF4FF}}c |}
\hline
\multicolumn{1}{|c|}{\cellcolor[HTML]{DAE8FC}Model} & Accuracy (\%)  \\ \hline
Baseline                                            & 70.25          \\ \hline
Baseline+Var                                        & 72.15          \\ \hline
Baseline+Doppler                                    & 72.82          \\ \hline
OURS                                                & \textbf{77.18} \\ \hline
\end{tabular}
\caption{Ablation experiments on the CSI-HAR dataset.}
\label{tab:ablation}
\end{table}

\begin{table*}[ht]
\begin{tabular}{|
>{\columncolor[HTML]{ECF4FF}}l |
>{\columncolor[HTML]{DAE8FC}}c 
>{\columncolor[HTML]{DAE8FC}}c 
>{\columncolor[HTML]{DAE8FC}}c 
>{\columncolor[HTML]{DAE8FC}}c 
>{\columncolor[HTML]{DAE8FC}}c |
>{\columncolor[HTML]{ECF4FF}}c 
>{\columncolor[HTML]{ECF4FF}}c 
>{\columncolor[HTML]{ECF4FF}}c 
>{\columncolor[HTML]{ECF4FF}}c 
>{\columncolor[HTML]{ECF4FF}}c |
>{\columncolor[HTML]{DAE8FC}}c 
>{\columncolor[HTML]{DAE8FC}}c 
>{\columncolor[HTML]{DAE8FC}}c 
>{\columncolor[HTML]{DAE8FC}}c 
>{\columncolor[HTML]{DAE8FC}}c |}
\hline
Window Length & \multicolumn{5}{c|}{\cellcolor[HTML]{DAE8FC}25}                                                                                                                                                                                                                                                                                              & \multicolumn{5}{c|}{\cellcolor[HTML]{ECF4FF}\textbf{50}}                                                                                                                                                                                                                                                                                              & \multicolumn{5}{c|}{\cellcolor[HTML]{DAE8FC}75}                                                                                                                                                                                                                                                                                              \\ \hline
Overlap (\%)  & \multicolumn{1}{c|}{\cellcolor[HTML]{DAE8FC}80}                           & \multicolumn{1}{c|}{\cellcolor[HTML]{DAE8FC}60}                           & \multicolumn{1}{c|}{\cellcolor[HTML]{DAE8FC}40}                           & \multicolumn{1}{c|}{\cellcolor[HTML]{DAE8FC}20}                           & 0                            & \multicolumn{1}{c|}{\cellcolor[HTML]{ECF4FF}80}                           & \multicolumn{1}{c|}{\cellcolor[HTML]{ECF4FF}60}                           & \multicolumn{1}{c|}{\cellcolor[HTML]{ECF4FF}40}                           & \multicolumn{1}{c|}{\cellcolor[HTML]{ECF4FF}\textbf{20}}                           & 0                            & \multicolumn{1}{c|}{\cellcolor[HTML]{DAE8FC}80}                           & \multicolumn{1}{c|}{\cellcolor[HTML]{DAE8FC}60}                           & \multicolumn{1}{c|}{\cellcolor[HTML]{DAE8FC}40}                           & \multicolumn{1}{c|}{\cellcolor[HTML]{DAE8FC}20}                           & 0                            \\ \hline
Accuracy (\%) & \multicolumn{1}{c|}{\cellcolor[HTML]{DAE8FC}{\color[HTML]{333333} 76.62}} & \multicolumn{1}{c|}{\cellcolor[HTML]{DAE8FC}{\color[HTML]{333333} 76.06}} & \multicolumn{1}{c|}{\cellcolor[HTML]{DAE8FC}{\color[HTML]{333333} 76.77}} & \multicolumn{1}{c|}{\cellcolor[HTML]{DAE8FC}{\color[HTML]{333333} 75.46}} & {\color[HTML]{333333} 76.39} & \multicolumn{1}{c|}{\cellcolor[HTML]{ECF4FF}{\color[HTML]{333333} 76.26}} & \multicolumn{1}{c|}{\cellcolor[HTML]{ECF4FF}{\color[HTML]{333333} 75.24}} & \multicolumn{1}{c|}{\cellcolor[HTML]{ECF4FF}{\color[HTML]{333333} 76.06}} & \multicolumn{1}{c|}{\cellcolor[HTML]{ECF4FF}{\color[HTML]{333333} \textbf{77.18}}} & {\color[HTML]{333333} 76.21} & \multicolumn{1}{c|}{\cellcolor[HTML]{DAE8FC}{\color[HTML]{333333} 75.65}} & \multicolumn{1}{c|}{\cellcolor[HTML]{DAE8FC}{\color[HTML]{333333} 76.58}} & \multicolumn{1}{c|}{\cellcolor[HTML]{DAE8FC}{\color[HTML]{333333} 76.06}} & \multicolumn{1}{c|}{\cellcolor[HTML]{DAE8FC}{\color[HTML]{333333} 75.91}} & {\color[HTML]{333333} 76.51} \\ \hline
\end{tabular}
\caption{Hyperparameter window length (N) and hop size (H) of equation \ref{eq:fourier} tuning for CSI-HAR dataset.}
\label{tab:window_length}
\end{table*}

We choose the hyperparameter window length (N) and hop size (H) of equation \ref{eq:fourier} empirically. Tab. \ref{tab:window_length} showcase the experimental results for the CSI-HAR dataset. We achieve the top performance with window length $50$ and hop length $40$ which represents $20\%$ overlap for this dataset.

\section{Attention Visualization}

\begin{figure}[!h]
    \centering
    \subfloat[Variance-based\label{fig:a}]{
        \includegraphics[width=0.8\linewidth]{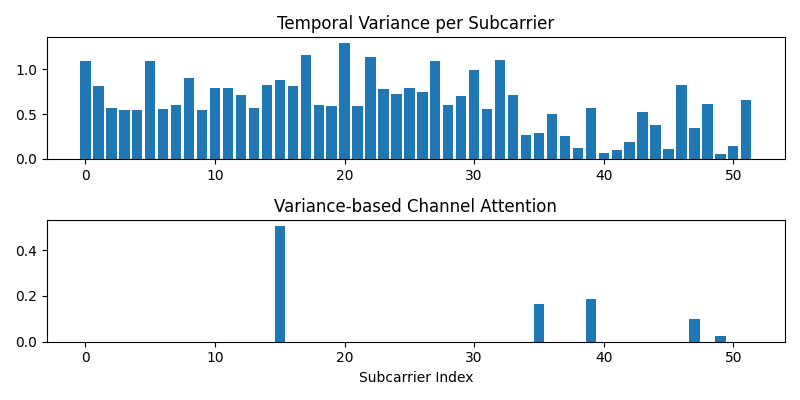}
        \label{fig:fall_variance}
    }
    \hfill
    \subfloat[Doppler-aware\label{fig:a}]{
        \includegraphics[width=0.8\linewidth]{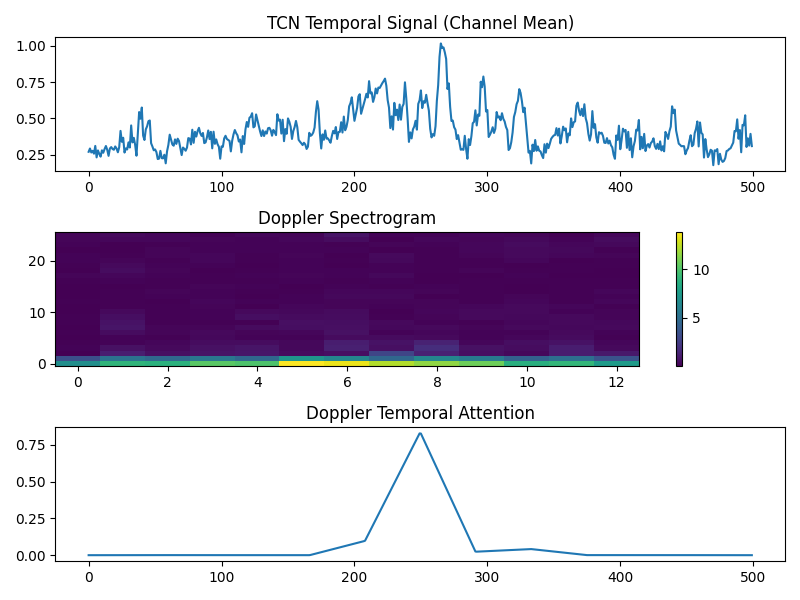}
        \label{fig:fall_doppler}
    }
    \caption{Attention visualization for a sample of the action class "bend" of the CSI-HAR dataset.}
    \label{fig:bend}
\end{figure}

\begin{figure}[!h]
    \centering
    \subfloat[Variance-based\label{fig:a}]{
        \includegraphics[width=0.8\linewidth]{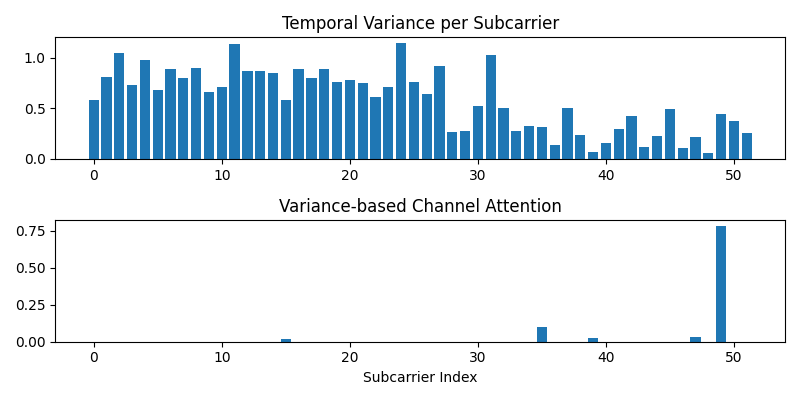}
        \label{fig:fall_variance}
    }
    \hfill
    \subfloat[Doppler-aware\label{fig:a}]{
        \includegraphics[width=0.8\linewidth]{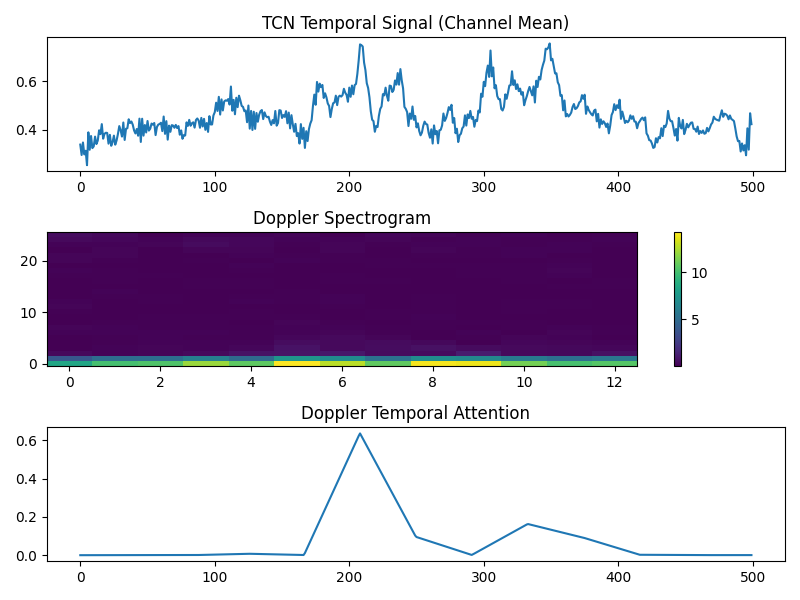}
        \label{fig:fall_doppler}
    }
    \caption{Attention visualization for a sample of the action class "fall" of the CSI-HAR dataset.}
    \label{fig:fall}
\end{figure}

\begin{figure}[!h]
    \centering
    \subfloat[\label{fig:a}]{
        \includegraphics[width=0.8\linewidth]{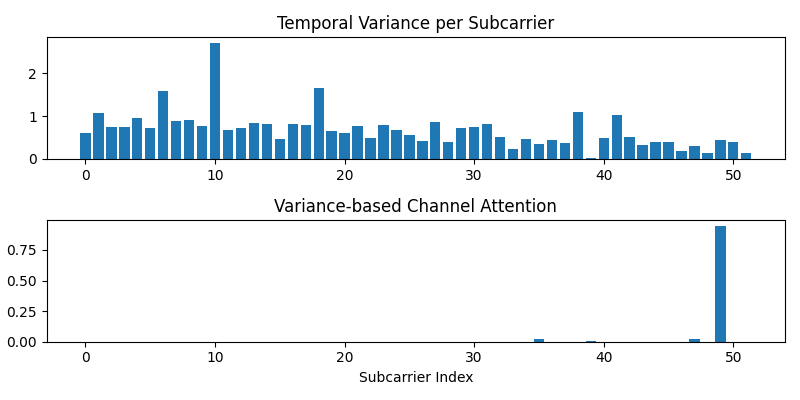}
        \label{fig:fall_variance}
    }
    \hfill
    \subfloat[\label{fig:a}]{
        \includegraphics[width=0.8\linewidth]{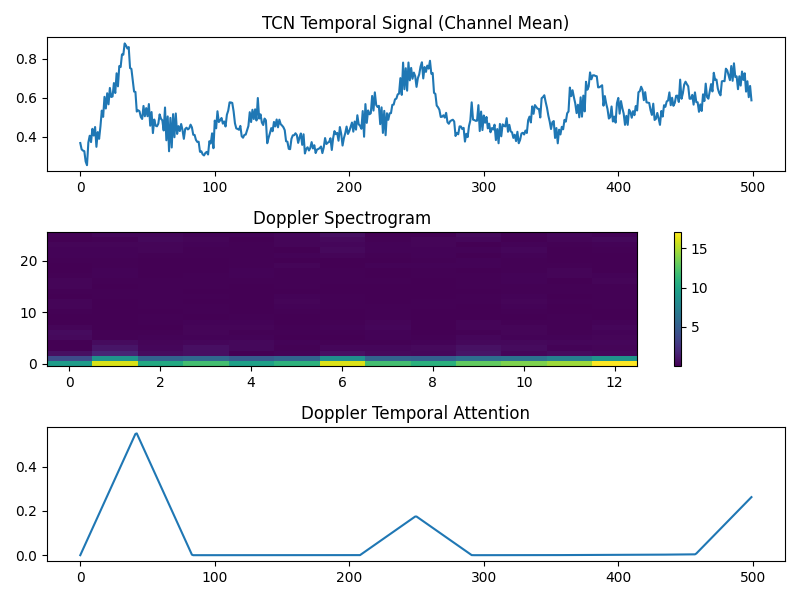}
        \label{fig:fall_doppler}
    }
    \caption{Attention visualization for a sample of the action class "walk" of the CSI-HAR dataset.}
    \label{fig:walk}
\end{figure}

\begin{figure}[!h]
    \centering
    \subfloat[\label{fig:a}]{
        \includegraphics[width=0.8\linewidth]{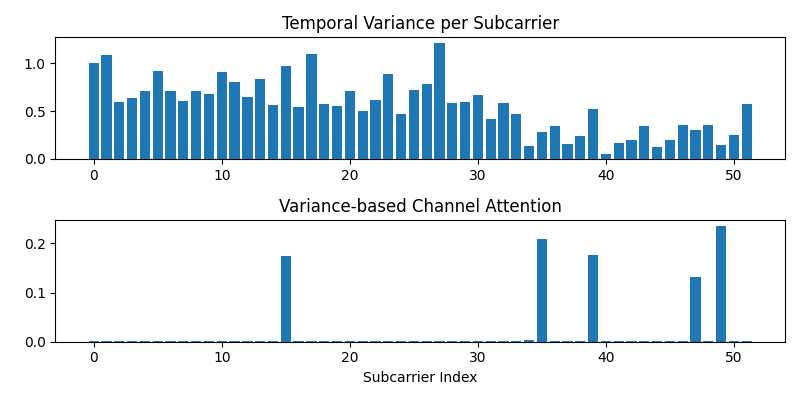}
        \label{fig:fall_variance}
    }
    \hfill
    \subfloat[\label{fig:a}]{
        \includegraphics[width=0.8\linewidth]{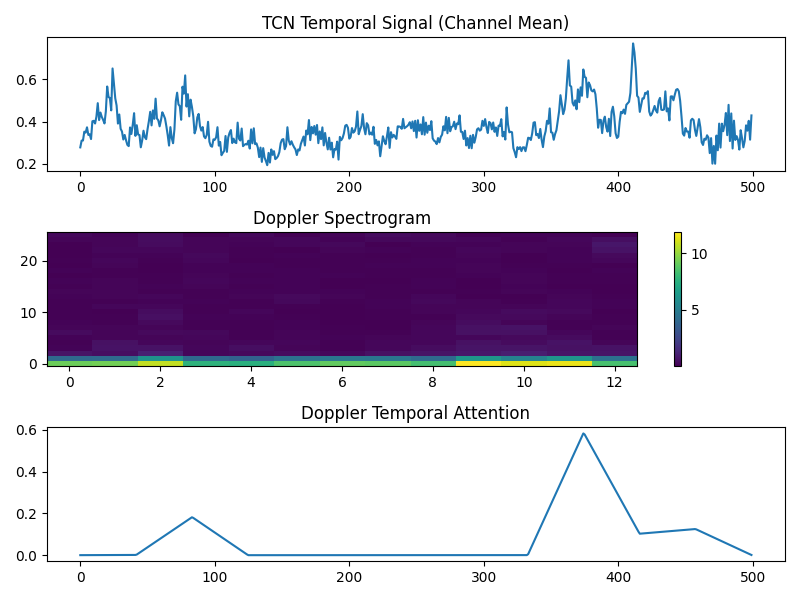}
        \label{fig:fall_doppler}
    }
    \caption{Attention visualization for a sample of the action class "run" of the CSI-HAR dataset.}
    \label{fig:run}
\end{figure}

\begin{figure}[!h]
    \centering
    \subfloat[\label{fig:a}]{
        \includegraphics[width=0.8\linewidth]{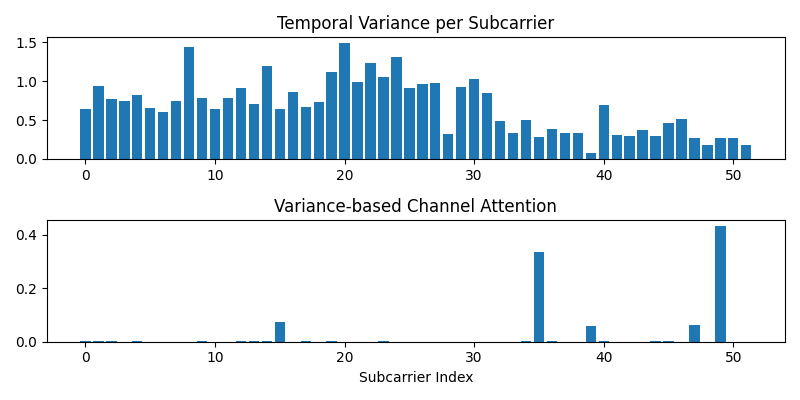}
        \label{fig:fall_variance}
    }
    \hfill
    \subfloat[\label{fig:a}]{
        \includegraphics[width=0.8\linewidth]{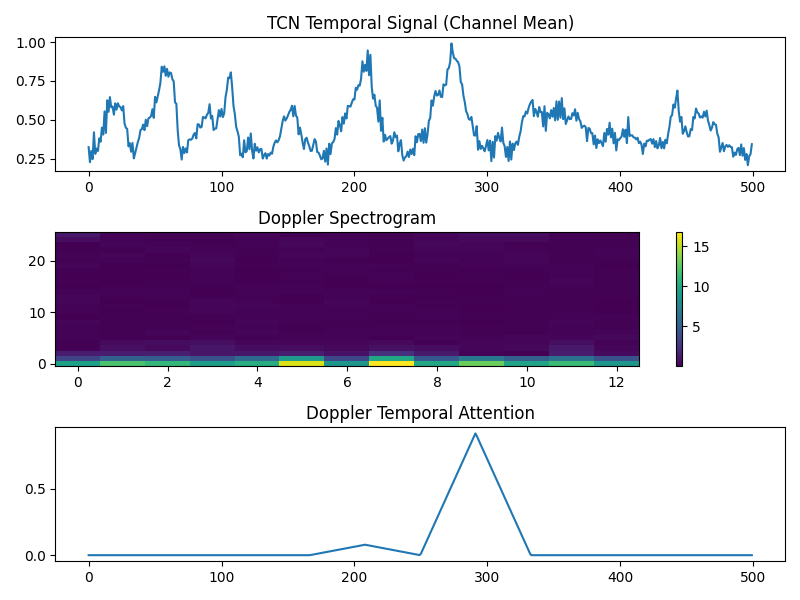}
        \label{fig:fall_doppler}
    }
    \caption{Attention visualization for a sample of the action class "lie-down" of the CSI-HAR dataset.}
    \label{fig:lie-down}
\end{figure}

\begin{figure}[!h]
    \centering
    \subfloat[\label{fig:a}]{
        \includegraphics[width=0.8\linewidth]{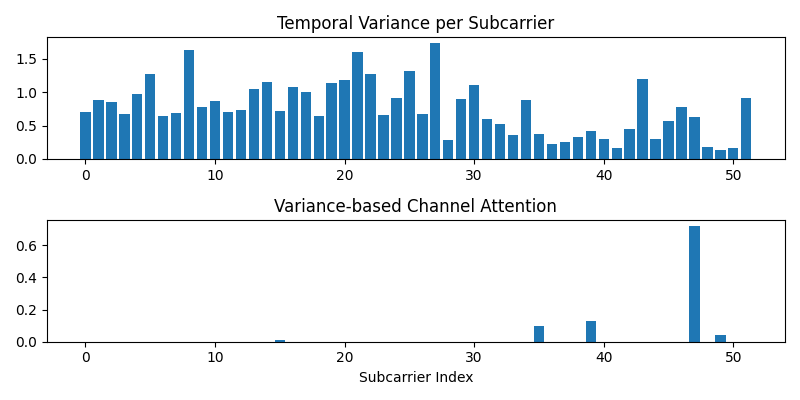}
        \label{fig:fall_variance}
    }
    \hfill
    \subfloat[\label{fig:a}]{
        \includegraphics[width=0.8\linewidth]{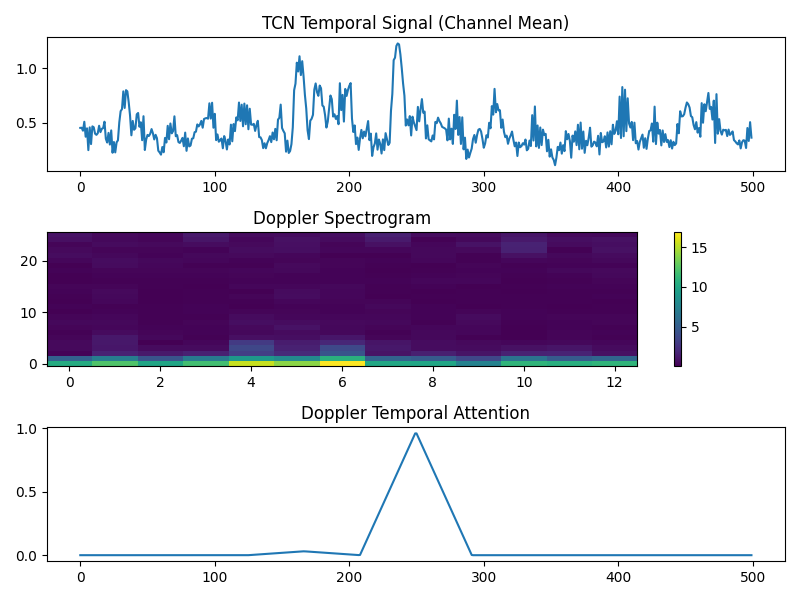}
        \label{fig:fall_doppler}
    }
    \caption{Attention visualization for a sample of the action class "stand-up" of the CSI-HAR dataset.}
    \label{fig:standup}
\end{figure}

\begin{figure}[!h]
    \centering
    \subfloat[\label{fig:a}]{
        \includegraphics[width=0.8\linewidth]{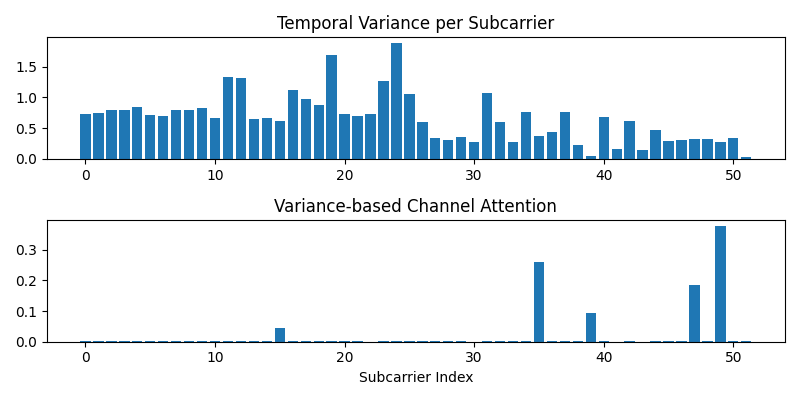}
        \label{fig:fall_variance}
    }
    \hfill
    \subfloat[\label{fig:a}]{
        \includegraphics[width=0.8\linewidth]{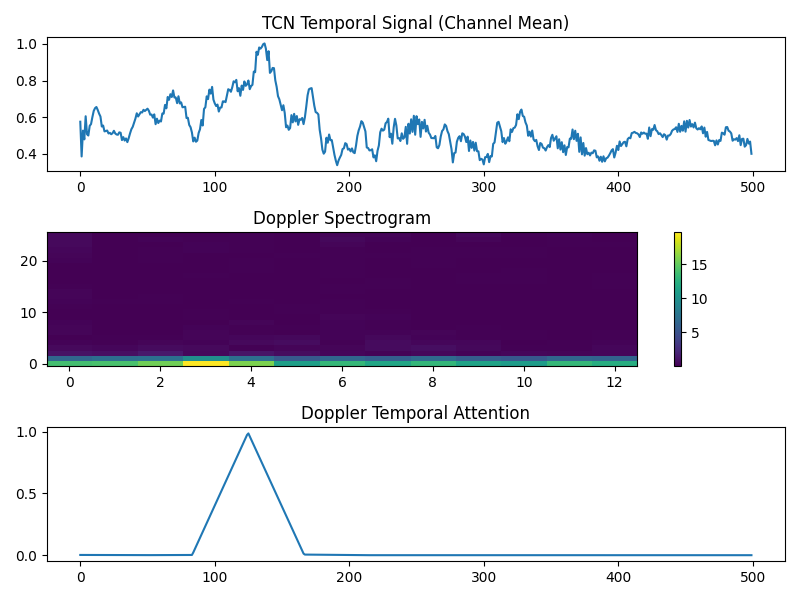}
        \label{fig:fall_doppler}
    }
    \caption{Attention visualization for a sample of the action class "sit-down" of the CSI-HAR dataset.}
    \label{fig:sitdown}
\end{figure}

Figs. \ref{fig:bend}–\ref{fig:sitdown} illustrate the learned attention behavior of the proposed framework for representative samples of the actions bend, fall, walk, run, lie-down, stand-up, and sit-down from the CSI-HAR dataset. In each figure, subfigure (a) shows the selected subcarriers along with their corresponding attention weights, while subfigure (b) depicts the temporal attention assigned to different time steps of the CSI signal. As observed, the proposed variance-based channel attention identifies informative CSI subcarriers by leveraging temporal variance, producing sparse attention distributions that emphasize discriminative channels while suppressing redundant ones. Concurrently, the Doppler-aware temporal attention captures motion dynamics through the Doppler spectrogram, assigning higher importance to time segments with elevated Doppler energy. By jointly exploiting channel-wise variance and Doppler-induced temporal dynamics, the model effectively focuses on the most relevant spatial and temporal features for accurate activity recognition.



\section{Conclusion}

This paper challenges the prevailing assumption that increasing model capacity is sufficient for WiFi CSI-based HAR. We show that the core limitation of existing approaches lies not in architectural depth, but in their failure to explicitly model motion dynamics inherent in CSI signals. To address this, we propose a compact TCN framework that embeds motion awareness directly into the feature space through two complementary mechanisms: Doppler-energy-based temporal attention, which selectively amplifies motion-salient temporal regions, and variance-driven channel attention, which prioritizes informative subcarriers based on temporal motion statistics.

Unlike conventional pipelines that rely on implicit feature learning from raw or transformed CSI, the proposed approach introduces an inductive bias aligned with the physical structure of the signal. This results in a model that is not only more parameter-efficient, but also more effective at capturing discriminative motion patterns. Extensive experiments demonstrate that explicit motion-aware modeling consistently outperforms deeper, computation-heavy baselines, while significantly reducing model complexity.

These findings suggest that future progress in CSI-based HAR should move beyond scaling architectures, and instead focus on integrating domain-specific motion priors into the learning process.


\bibliographystyle{IEEEtran}
\bibliography{bare_jrnl_new_sample4}
\vfill

\end{document}